%% file: paper.tex
\documentclass[10pt, conference, compsocconf]{IEEEtran}
\IEEEoverridecommandlockouts 

\usepackage{cite}

\usepackage[hyphens,spaces,obeyspaces]{url}
\urldef{\footurl}\url{https://github.com/JWijkhuizen/model_training}

\usepackage{amsmath,amssymb,amsfonts}
\usepackage{algorithmic}
\usepackage{graphicx}
\usepackage{textcomp}
\usepackage[table,xcdraw]{xcolor}
\usepackage{dblfloatfix}

\usepackage{caption}
\usepackage{subcaption} 
\usepackage{framed}
\usepackage{listings}
\usepackage{makecell}
\usepackage{array, booktabs}
\usepackage[
  capitalize
]{cleveref}

\begin{document}

\title{Context-based navigation for ground mobile robot in semi-structured indoor environment }

\author{\IEEEauthorblockN{Darko Bozhinoski}
\IEEEauthorblockA{IRIDIA, University Libre du Bruxelles \\
 Email: darko.bozhinoski@ulb.be}
\and
\IEEEauthorblockN{Jasper Wijkhuizen}
\IEEEauthorblockA{Cognitive Robotics Department, TU Delft \\
 Email: j.wijkhuizen@gmail.com}}
\maketitle

\begin{abstract}

There is a growing demand for mobile robots to operate in more variable environments, where guaranteeing safe robot navigation is a priority, in addition to time performance. To achieve this, current solutions for local planning use a specific configuration tuned to the characteristics of the application environment. 
In this paper, we present an approach for developing quality models that can be used by a self-adaptation framework to adapt the local planner configuration at run-time based on the perceived environment. We contribute a definition of safety model that predicts the safety of a navigation configuration given the perceived environment.
Experiments have been performed in a realistic navigation scenario for a retail application to validate the obtained models and demonstrate their integration in a self-adaptation framework.
\end{abstract}
\begin{IEEEkeywords}
Autonomous Robot Navigation, Software Architecture for Robotic Systems, Self-Adaptive Systems, Motion and Path Planning, Component-based engineering
\end{IEEEkeywords}

\input{Sections/Introduction}

\input{Sections/Related_work}

\input{Sections/Main}
\input{Sections/Evaluation}

\input{Sections/Discussion}
\input{Sections/Conclusion}

\bibliographystyle{IEEEtran}
\bibliography{IEEEabrv, references}




\end{document}

%% file: Sections/Introduction.tex
\section{Introduction}

Mobile robots are used in many industrial applications to increment productivity (e.g. in warehouses, factories).
However, there is a growing demand for mobile robots to operate in complex environments with increasing environment variability. Retail stores are environments which are semi-structured\textemdash obstacles such as boxes, crates and other obstacles can appear in new places that are unknown apriori to the robot. The area where the robot navigates is shared with people, which makes navigation extremely challenging 
since the movement of humans can be highly unpredictable. In this context, ensuring safety of the humans and the robot itself is a priority, in addition to the required performance of the robot navigation to improve general productivity.

In general, the quality with which a robot performs its mission is multi-dimensional and can be defined through different aspects or \emph{quality attributes} of the robot system, including safety, performance, reliability and robustness \cite{Berrocal2018RoQME:Metrics}. Some of these quality attributes conflict with each other and a trade-off is necessary. (e.g. safety and performance).


In relation to safety, the key component of robot navigation architectures is the local path planner that is responsible for obstacle avoidance \cite{Siegwart2004IntroductionRobots}.
Different planning algorithms have been proposed to address
dynamic and partially known environments, 
which typically require to be optimized for a specific scenario and environment type \cite{Mac2016HeuristicSurvey}, and achieve different levels of performance and safety. 
There is no algorithm configuration optimal for every environment.
Hence, navigation in environments whose characteristics vary during the robot operation is still an open research issue\cite{Siegwart2004IntroductionRobots}.
Different robot configurations might favor certain quality attributes, so choosing an "optimal" configuration for a specific situation requires that the trade-off between the quality aspects is explored \cite{Brugali2019Non-functionalArt}. The choice of which configuration of a local planner is deployed is done by an engineer at design-time. However, if the environment varies during operation then the deployed configuration needs to change. 
The demand for a higher autonomy in robotic systems requires that robots are able to make a decision and switch between available alternative configurations 
\textemdash acting as a self-adaptive system. Self-adaptive systems are systems that are able to change their configurations during mission execution depending on the context where they operate \cite{Brugali2019Non-functionalArt}.
In this paper, we define navigation quality of a mobile robot operating in diverse environments in terms of safety and performance. We present an approach for building \emph{quality models} that can: (i) identify when the robot should change configuration; (ii) decide which configuration is optimal for a specific environment.
We apply our constructed models to a self-adaptation solution in a retail navigation scenario and compare it to a baseline system in terms of performance and safety.

We make the following contributions:
\begin{itemize}
  \item Definition of metrics for safety and perfomance in robot navigation scenarios.
  \item Definition of narrowness and obstacle density metrics to characterize the robot environment.
  \item A method to build quality models that correlate safety with the environment metrics.
  \item Application of the quality model in a self-adaptation framework to adapt the local planner configuration based on the environment at runtime 
\end{itemize}

The rest of the paper is organized as follows. Related work is examined in Section~\ref{RW}. Section~\ref{ContextN}    describes the main contribution of our approach.
Section~\ref{Navigtionmetric} defines the metrics for safety and performance  and the environmental metrics that influence them in robot navigation scenarios. Section~\ref{ConstructingModels} presents the constructed safety model which predicts the quality of a configuration based on environmental metrics. Section~\ref{Evaluation} presents the experiments that analyze the usefulness of quality models for a robotics self-adaptation framework. Finally, Section~\ref{Conclusion} concludes the paper.

%% file: Sections/Related_work.tex
\section{Related work}\label{RW}
At present, solutions  for  local  planning  use  a  specific parametrization of the algorithm tuned  to  the  characteristics of   the   application   environment.
\cite{Kulich2015} compared three of the most used local planning algorithms:  Smooth Nearness-Diagram(SND), Dynamic Window Approach (DWA), and Vector Field Histogram (plus)(VFH+). The results showed that SND is better than the DWA for narrow environments, while DWA and the VFH+ method outperform SND in wide environments. In another study, \cite{Sanchez2018ImplementationVehicles} showed that  Timed Elastic Bands(TEB) performs better than DWA in an environment with narrow corridors. 

Different metrics for robot navigation quality have been evaluated in the literature in relation to safety and performance, such as mean distance between robot and obstacles, distance covered by the robot, time to completion, and smoothness of the trajectory \cite{Munoz2007EvaluationRobot}. 
Many of these metrics are available only after mission completion. Here, we define safety and performance metrics that can be computed at run-time.

   In terms of self-adaptation in robotics, most works \cite{Brugali2018, Gherardi2015}, focus on frameworks that define rules for adaptation of generic robotic solutions. In this work, we develop machine learning models using preexisting local planner configurations. Our research
  enriches the knowledge of adaptation frameworks to make more informed decisions.

%% file: Sections/Main.tex
\section{Context-based navigation}\label{ContextN}

We present an approach based on a  self-adaptation framework that adapts the local planner configuration at run-time based on the specificity of the  perceived  environment. To implement the self-adaptation capability, we use the MROS model-based framework\cite{HernandezCorbato2020MROS:Architectures} that adapts robot control architectures to satisfy set of quality requirements. We choose MROS because it is a model-based solution that integrates easily in a typical ROS development workflow and promotes reusability.  
To decide for an adaptation, MROS  uses the (functional) knowledge of the robot system configurations together with the perceived quality levels of the current configuration. This research aims to enrich the knowledge base of the MROS framework with additional  source of information about the environment where the robot is performing the mission. We created a link between the different environments and the perceived robot safety level through a machine learning model. 

 

We contribute two novel elements that enable runtime adaptation to changes in the environment:
\begin{enumerate}
    \item Characterization of the navigation application model, in terms of metrics for safety and performance, the characteristics of the mission environment that have the most influence in these quality attributes, and the model of the navigation architecture. 
    \item Navigation quality model based on environment metrics using machine learning for each configuration. \\
In this work, we constructed a safety model for local planner configurations. Simulations in different environments are conducted and data is collected by measuring the navigation quality levels and the environment characteristics each time-step. Different train and test data-sets are constructed to train and cross-validate the obtained quality models.


  

     
\end{enumerate}


\subsection{MROS Framework} 
\noindent
 MROS implements a MAPE-K loop using a runtime model conforming to the TOMASys meta-model serving as a knowledge base (KB) \cite{HernandezCorbato2020MROS:Architectures}. Here, we summarize the main phases in the MROS MAPE-K loop.  

\textit{Monitor}. Two monitoring processes track the status of active ROS nodes and the degree of fulfillment of non-functional requirements using quality attributes as a proxy.  The goal is to: 
(i) identify erroneous node behavior; and (ii) observe violation of system constraints (i.e. level of quality attributes). The monitors operate at a fixed frequency and produce diagnostics using the standard ROS diagnostic mechanism. 
\emph{ROS node monitors} publish a diagnostic error message in case a functional error is detected.  \emph{Quality attribute monitors} are implemented by a separate \emph{observer} node per each quality attribute. 
Developers need to implement the logic in the observers to obtain the quality attribute level from the given signals. 

\textit{Analyze}. The system diagnostic data produced in the Monitor phase is used to infer the status of the functional architecture of the system.  Any new facts, for instance a new component status when a node has reported a fatal error, or a change in the level of a quality attribute, are asserted in the TOMASys model (KB). Then, ontological reasoning and application-independent rules (declared in SWRL \cite{HernandezCorbato2020MROS:Architectures}) are applied to the model to infer the current status of the functional architecture.
\looseness=-1

\textit{Plan}. If the meta-controller discovers that any objectives are violated, an architectural adaptation is proposed by searching for alternative function designs in the TOMASys model (KB). If several available designs meet the required quality attribute levels, the one maximizing an application-specific utility function is selected. 
\looseness=-1

\textit{Execute}. Enforce the adaptation selected in the Plan phase.

\section{Navigation application model}\label{Navigtionmetric}
In this section, we discuss the model developed to characterize a navigation application in terms of the navigation quality and the relevant characteristics of the mission environment.
\label{nav_quality_approach}
   We define navigation quality as a two-dimensional concept of performance and safety.
We develop two metrics:
 
    


    \subsubsection{Safety metric}
    We represent safety through the concept of a safety margin around a mobile robot  \cite{Chung2009SafeEnvironment}. 
    The safety level during mission execution is determined by the distance to obstacles: obstacles far from the robot do not influence safety. 
    The size of the safety margin is determined using the concept of braking distance, which is dependent on the current forward velocity of the robot and the maximum deceleration as noted in (\ref{Eq:Safety}). 
        \begin{equation}
        \label{Eq:Safety}
            d_{brake} = \frac{v_x^2}{2 a_{max}} 
        \end{equation}
        where  $a_{max}$ is a constant, defined by the robot's limits\textemdash taking into account robot's width and length. 
        Since the braking distance depends on the robot velocity, the safety margin is also velocity dependent. 
        We define safety as an indicator of the collision risk the robot perceives and we calculate it at runtime as follows:
        
        \begin{equation}\label{eq:S1}
            \text{Safety} =
            \begin{cases}
                1       & \quad \text{if } d_{obs} \geq d_{brake}\\
                \frac{d_{obs}}{d_{brake}}  & \quad \text{if } d_{obs} < d_{brake}
            \end{cases}
        \end{equation}
        where $d_{obs}$ is the distance to the nearest obstacle.
(\ref{eq:S1}) states that when there are no obstacles inside the safety margin, safety is 1. When the robot hits an obstacle 
safety is equal to 0. The robot safety level scales linearly based on the distance to the closest obstacle.
        
    \subsubsection{Performance metric}
\begin{figure}[h!]
            \centering
               \vspace{-0.45cm}
            \includegraphics[width=0.7\linewidth]{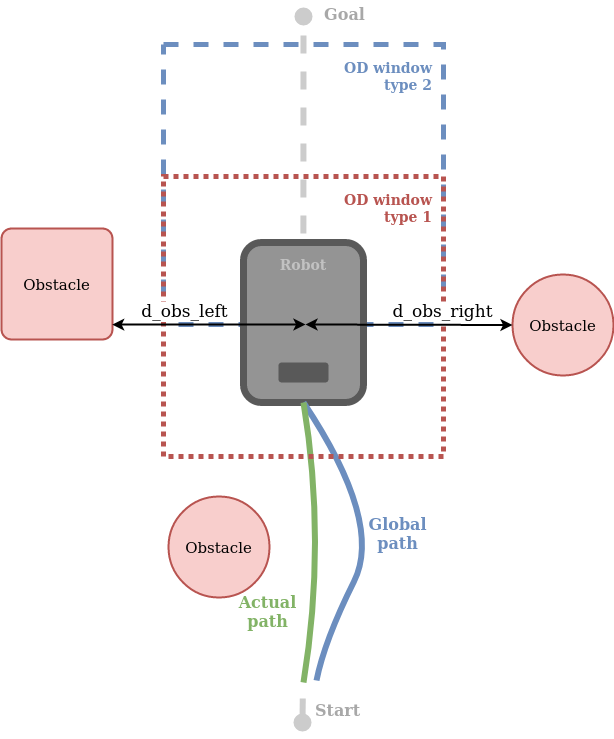}
            \caption{Visual explanation for the reference path, the Obstacle Density window and Narrowness}
            \label{fig:metrics}
            \vspace{-0.45cm}
        \end{figure}

           We define a performance metric based on the time to completion and the deviation from the global path from \cite{Steinfeld2006CommonInteraction}. This metric uses two seconds of the actual path, and compares it to the time to completion of a reference path. By using information from two past seconds and not the complete run, the performance level can be evaluated at run-time (after two seconds). We choose a time segment of 2 seconds because the performance measurements for less than 2 seconds did not provide enough information for a good variability in the signal. 
           The time to completion of the reference path is calculated using the path length and the maximum allowed speed for the robot as shown in (\ref{eq:P33}).
            \begin{equation}\label{eq:P33}
                t_{ref} = d_{ref} / v_{max}
            \end{equation}
            Finally, we define a \textit{performance metric} as the ratio between the time to completion of the actual path and the reference path:
            \begin{equation}\label{eq:P3}
                P = t_{ref}/t_{act} 
            \end{equation}
            where $t_{act}$ is a constant at $t_{act} = 2 s$.
         To obtain the reference path, the last two seconds of the  actual path of the robot is projected on its global path (See Figure~\ref{fig:metrics}).

\subsection{Environment Metrics}
Given a certain local planner configuration to operate in different environments result in  different levels of safety and performance.
To reason about safety and performance, the robot needs to be able to identify the environment where it performs its mission. We define two relevant environment metrics:  \textit{Obstacle Density} and \textit{Narrowness}. We consider that these two environment characteristics can be valuable information to define navigation quality due to the fact that: (i) they have
been previously discussed in one form or another in an experimental setup for a navigation of a mobile robot; and (ii) they can be measured using information from the robots' sensors. 
    

 
    \subsubsection{Narrowness}
       The concept of narrowness is inspired by the work of \cite{Kulich2015}. In this study, the authors state that DWA shows good results when the environment is wider than twice the diameter of the robot. This indicates that the free space distance, which is the distance to the closest obstacle, on both sides of the robot ($d_{obs\_left}$, $d_{obs\_right}$)
       is extremely relevant for the navigation quality of the local planner. 
    We define \textbf{narrowness} as a metric that indicates how many robot widths fit in the width of the current environment as shown in (\ref{eg:narrowness}). 

        \begin{equation}\label{eg:narrowness}
            N = \frac{d_{obs\_left} + d_{obs\_right}}{r_{width}}
        \end{equation}
where $d\_obs = min(range\_max, d\_obs)$.
The environment width is measured from the center of the robot, in the direction perpendicular to the direction of travel. The free space distance, 
is determined using the lidar data (see Figure \ref{fig:metrics}). For engineers to be able to use this metric for any mobile robot, regardless its size, the final value is obtained by dividing it to the width of the robot ($r_{width}$). A maximum range of the visibility space is considered since obstacles far away from the robot do not influence safety and performance. 
We 
explored different variations of the maximum range starting from 1m to 4m. We concluded that narrowness with max range
of 1m has the highest correlation with safety, and is therefore the best window to use when building a safety model. The correlation between narrowness and performance was inconclusive.

        
    \subsubsection{Obstacle density}
Information about the amount of obstacles around the robot have been used as parameters in experimental studies of a mobile robot in a navigation scenario by 
\cite{HernandezCorbato2020MROS:Architectures}.
While the number of obstacles is a relevant parameter, it does not convey information about the placement of these obstacles in the surrounding of the robot. Hence, we define \textbf{Obstacle Density (OD)} as a metric that indicates the ratio of the  occupied area in a local window around the robot:

       \vspace{-0.4cm} 
        \begin{equation}\label{eq:OD}
            OD = \text{Occupied area} / \text{Total area} 
                  \vspace{-0.2cm} 
        \end{equation}

        
        As robots are moving in particular direction, we hypothesize that the obstacles that have still not been encountered by the robot in the direction of moving are more relevant for the navigation quality than the obstacles to which the robot have come across in the past.  
        We consider two types of windows: (i) window centered around the robot, and (ii) window where the center is shifted to the front of the robot such that the window's rear side intersects with the robot's center (see Figure \ref{fig:metrics}).

           In ROS navigation, the information about the occupied area is taken from the local costmap. 
           The default size for the local costmap in ROS is 5m x 5m. To evaluate which size of the local window has the most influence on safety and performance we explored sizes equal or smaller than the default: 4mx4m, 3mx3m, 2mx2m, 1mx1m. The results showed that the correlation between safety and a small window is higher i.e. using a window of 1x1m is more advantageous when building a safety model. The correlation between obstacle density and performance was inconclusive.
        



\section{Building quality models}\label{ConstructingModels}
The robot should be able to estimate the quality with which a local planner configuration will navigate in a certain environment. Here, we present an approach to develop a safety model which predicts the safety quality of a configuration based on environmental metrics. It is important to note that  because the correlation between the environment metric and the performance were inconclusive we did not develop a performance predictive model. Instead, the performance for each configuration is obtained through the time to completion on a set of mission runs. These values are normalized and considered for reasoning on the metacontrol level. 

\subsection{Navigation architecture model}\label{navConfiguration}
For the navigation architecture we chose two of the most popular local planners in the ROS Navigation Stack: Dynamic Window Approach (DWA) and the Timed Elastic Bands
 (TEB). These methods were chosen  due to the fact that earlier studies \cite{Kulich2015}\cite{Sanchez2018ImplementationVehicles} show that they show different performance in narrow, wide, and cluttered environments. 
    We analysed the parameters of DWA and TEB and how they might influence the quality metrics.  Based on the initial tests we selected 3 configuration parameters per local planner. We chose the maximum speed, the sim\_time and the scaling\_speed for the DWA. We selected the maximum speed, the inflation\_radius and the weight\_obstacle for the TEB. We defined 16 different configurations: 8 per local planner. 
\subsection{Quality model construction}\label{env1}
We trained a separate safety model for each selected configuration. 
We use 
Linear Regression to train a safety model,  
because it is suited for a low number of features and a single output. We use narowness and obstacle density as input variables, while the output is the runtime safety level. To build the models we collected data from different set of environments, using a floorplan of a retailer in Western Europe.
We generated variations of environments for a simple corridor
 with different corridor width and clutterness. 

   \begin{figure*}[htp]
    \centering
    \includegraphics[width=0.7\textwidth]{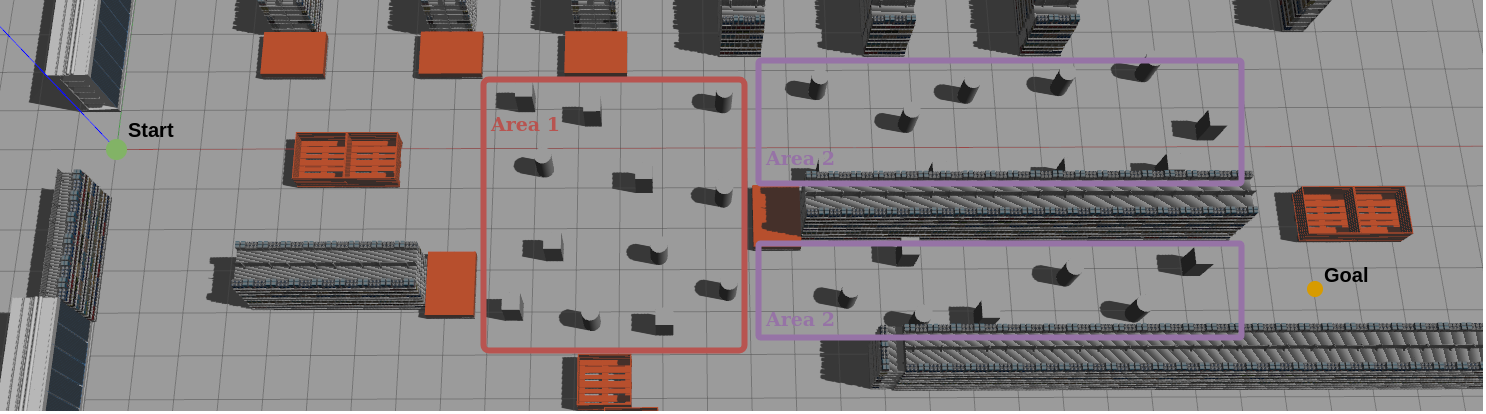}
    \caption{Simulation environment for the experiments. Obstacles are randomly spawned  in Area1 \&\& Area2.}
    \label{fig:final_exp_env}
      \vspace{-0.5cm}
\end{figure*}

   The models were tested on a mission where the robot moves from the beginning of a corridor to the end of the corridor with obstacles at random locations inside the corridor.
  We generated 5 environment instances for each environment type (4 different environment types), resulting in twenty different data-sets. 
  The twenty data-sets are split in training data (16 data-sets) and test data (4 sets). 
  The model for each configuration is trained five times, and the model with the highest score on its test set is selected.
  Each single run generated 300+ datapoints. For the training of a model of a single configuration we used four of the datasets, resulting in 4.8k+ data points for training (a training set contains 16 runs in total). The models were scored on the test sets using the coefficient of determination and the mean squared error.
    We performed validation runs in the retail store shown in Figure \ref{fig:final_exp_env}. Here, the robot does not drive only through a simple corridor, but in the whole supermarket. In the store, 
    the average coefficient of determination for all quality models is 0.85 (min=0.74, max=0.92), and the average mean squared error is 0.0071 (min=0.0045, max=0.0114). 
    The results show strong correlation between the environment metrics and the safety level, indicating that the developed safety models can be used for reasoning at runtime.

%% file: Sections/Evaluation.tex
\section{Experiments}\label{Evaluation}
\subsection{Experimental Setup}
\begin{figure*}[b]
\centering
 \hspace*{-2cm}
\advance\rightskip-1cm
      \begin{subfigure}[t]{0.4\textwidth}
              \centering
              \includegraphics[width=\textwidth]{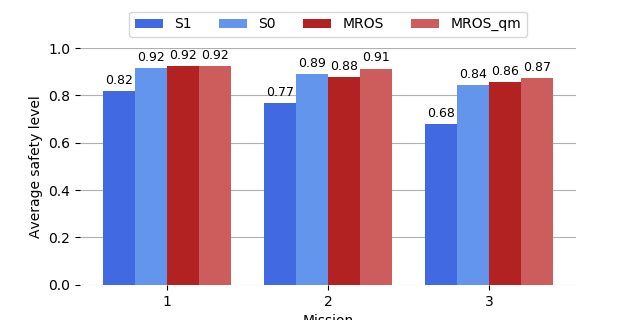}
              \caption{Average safety levels}
              \label{fig:barplot_safety_av}
        \end{subfigure}
        \hfill
         \hspace*{-2cm}
\advance\rightskip-1cm
    \begin{subfigure}[t]{0.4\textwidth}
      \centering
          \includegraphics[width=\textwidth]{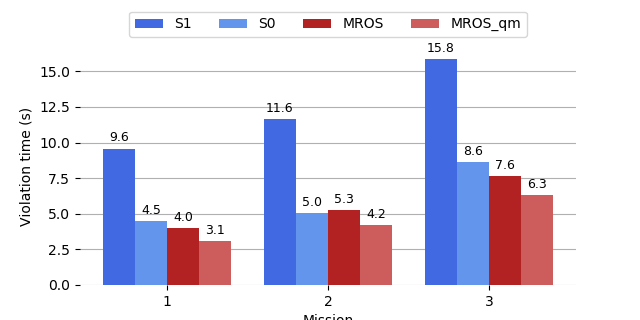}
          \caption{Time under a safety threshold}
          \label{fig:barplot_violate}
    \end{subfigure}
    \begin{subfigure}[t]{0.4\textwidth}
      \centering
          \centering
        
          \includegraphics[width=\textwidth]{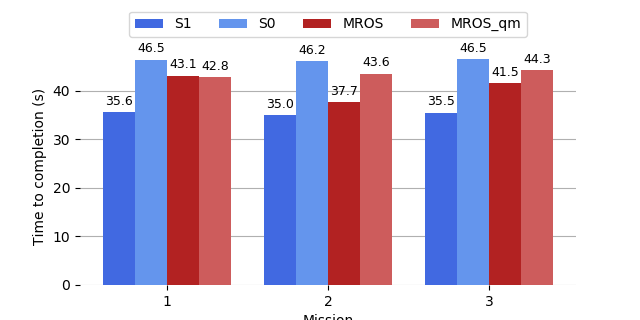}
          \caption{Time to completion}
          \label{fig:barpot_timetc}
    \end{subfigure}
    \caption{Experimental Results.}
\end{figure*}

We  analyze the usefulness of quality models for a robotics self-adaptation framework with respect to the improvement in the overall navigation quality. 
We compared 4 different systems:
(i) MROS\_qm is a setup of a robotic system that uses the metacontrol framework combined with the obtained quality models; (ii) MROS is a robotic system that uses a metacontrol framework without quality models; (iii) S0 is a setup that uses the "safest" configuration; (iv) S1 is a setup that uses the "fastest" configuration.  

     
         %

To monitor the context and runtime execution,  we created 4 MROS observers: 2 observers to monitor the quality attributes (safety and performance) and two to monitor the environmental context (obstacle density and narrowness) on the current configuration. 





   The experiments were run on a navigation mission. We specified 3 mission types that differ in their obstacle density and the clutterness of Area1 and Area2 (Figure~\ref{fig:final_exp_env}). 
   We measured the following dependent variables: (i) time to completion (performance); (ii) average safety level over a complete run; (iii) time under a safety threshold (the threshold corresponds to a safety constraint with a value of 0.6). 

\textbf{Replication package.} A replication package of this work is publicly available to interested researchers. The replication package\footnote{\footurl} consists of two parts: (i) scripts to train the models  and (ii) scripts to run the experiments and the raw data obtained from the experiments. The floorplan of the store is not provided because it is a proprietary confidential information.

\subsection{Experimental Results and Discussion}





 Figure~\ref{fig:barplot_safety_av} presents the average safety levels for all systems. The results indicate that  MROS\_qm  has highest average safety level for every mission. It is interesting to note that MROS\_qm  outperforms both: MROS and S0 - the system with the safest average configuration. 
 In MROS\_qm, the reasoner knows which configurations will violate the safety constraint in the current environment, and therefore makes a more informed decision for adaptation. MROS cannot predict which configuration will violate the safety constraint and selects the next best available configuration based on average safety and performance levels. 
 
 Figure~\ref{fig:barplot_violate} shows the time that a system violates the safety constraint. The results indicate that MROS\_qm outperforms the other systems in all 3 missions (labeled 1, 2 and 3). As expected S1 shows the worst results: the time it violates the safety constraint is 2.5 - 3 times more than  MROS\_qm. 
 
 Figure~\ref{fig:barpot_timetc} shows the time a system completes a mission. MROS\_qm shows lower performance than MROS in two of the three missions since more often it favors a safer configuration rather than a faster one. However, it is interesting to note that MROS\_qm  completes the mission faster than MROS for Mission 1 and outperforms S0 for all missions. As expected, S1 performs the mission the fastest.

%% file: Sections/Conclusion.tex
\section{Conclusion}\label{Conclusion}
   
The aim of this research is to establish methods for creating an advanced reasoning knowledge base for self-adaptive robot systems. In this work, 
we propose a method to build safety quality model to be used in a self-adaptation framework. 

Our future work is in the following directions: (i) extend the definition of navigation quality through other quality attributes; (ii) expand the scope of machine learning methods used to develop quality models; and (iii) apply the constructed quality models in a proactive self-adaptation framework that performs adaptation before the system constraints are violated.
      
\section{Acknowledgements}This work is partially supported by Ahold Delhaize.
Darko Bozhinoski acknowledges support from the Belgian Fonds de la Recherche Scientifique–FNRS [No:40001145].


%% file: paper.bbl
\begin{thebibliography}{10}
\providecommand{\url}[1]{#1}
\csname url@samestyle\endcsname
\providecommand{\newblock}{\relax}
\providecommand{\bibinfo}[2]{#2}
\providecommand{\BIBentrySTDinterwordspacing}{\spaceskip=0pt\relax}
\providecommand{\BIBentryALTinterwordstretchfactor}{4}
\providecommand{\BIBentryALTinterwordspacing}{\spaceskip=\fontdimen2\font plus
\BIBentryALTinterwordstretchfactor\fontdimen3\font minus
  \fontdimen4\font\relax}
\providecommand{\BIBforeignlanguage}[2]{{%
\expandafter\ifx\csname l@#1\endcsname\relax
\typeout{** WARNING: IEEEtran.bst: No hyphenation pattern has been}%
\typeout{** loaded for the language `#1'. Using the pattern for}%
\typeout{** the default language instead.}%
\else
\language=\csname l@#1\endcsname
\fi
#2}}
\providecommand{\BIBdecl}{\relax}
\BIBdecl

\bibitem{Berrocal2018RoQME:Metrics}
\BIBentryALTinterwordspacing
J.~Berrocal, J.~Garcia-Alonso, and J.~Hern{\'{a}}ndez, ``{RoQME: Dealing with
  Non-Functional Properties through Global Robot QoS Metrics},'' Tech. Rep.,
  2018. [Online]. Available: \url{www.uidee.com}
\BIBentrySTDinterwordspacing

\bibitem{Siegwart2004IntroductionRobots}
R.~Siegwart and I.~R. Nourbakhsh, \emph{{Introduction to Autonomous Mobile
  Robots}}, 2004, vol.~2, no.~1.

\bibitem{Mac2016HeuristicSurvey}
T.~T. Mac, C.~Copot, D.~T. Tran, and R.~De~Keyser, ``{Heuristic approaches in
  robot path planning: A survey},'' \emph{Robotics and Autonomous Systems},
  2016.

\bibitem{Brugali2019Non-functionalArt}
D.~Brugali, ``{Non-functional requirements in robotic systems: Challenges and
  state of the art},'' in \emph{2019 IEEE International Conference on Real-Time
  Computing and Robotics, RCAR 2019}, vol. 2019-Augus.\hskip 1em plus 0.5em
  minus 0.4em\relax Institute of Electrical and Electronics Engineers Inc., 8
  2019, pp. 743--748.

\bibitem{Kulich2015}
M.~Kulich, V.~Koz{\'{a}}k, and L.~P{\v{r}}eu{\v{c}}il, ``{Comparison of local
  planning algorithms for mobile robots},'' in \emph{Lecture Notes in Computer
  Science}, vol. 9055.\hskip 1em plus 0.5em minus 0.4em\relax Springer Verlag,
  2015, pp. 196--208.

\bibitem{Sanchez2018ImplementationVehicles}
J.~Sanchez, ``{Implementation and comparison in local planners for Ackermann
  vehicles},'' 2018.

\bibitem{Munoz2007EvaluationRobot}
N.~D. Mu{\~{n}}oz, J.~A. Valencia, and N.~Londo{\~{n}}o, ``{Evaluation of
  navigation of an autonomous mobile robot},'' \emph{Performance Metrics for
  Intelligent Systems (PerMIS) Workshop}, vol.~1, pp. 15--21, 2007.

\bibitem{Brugali2018}
D.~Brugali, R.~Capilla, R.~Mirandola, and C.~Trubiani, ``{Model-based
  development of QoS-aware reconfigurable autonomous robotic systems},'' in
  \emph{IEEE International Conference on Robotic Computing, IRC 2018}, vol.
  2018-Janua, 4 2018, pp. 129--136.

\bibitem{Gherardi2015}
L.~Gherardi and N.~Hochgeschwender, ``{RRA: Models and tools for robotics
  run-time adaptation},'' in \emph{IEEE International Conference on Intelligent
  Robots and Systems}, vol. 2015-Decem.\hskip 1em plus 0.5em minus 0.4em\relax
  Institute of Electrical and Electronics Engineers Inc., 12 2015, pp.
  1777--1784.

\bibitem{HernandezCorbato2020MROS:Architectures}
C.~Hernandez~Corbato and D.~Bozhinoski, ``{MROS: Runtime adaptation for robot
  control architectures},'' 2020.

\bibitem{Chung2009SafeEnvironment}
W.~Chung, S.~Kim, M.~Choi, J.~Choi, H.~Kim, C.~B. Moon, and J.~B. Song, ``{Safe
  navigation of a mobile robot considering visibility of environment},''
  \emph{IEEE Transactions on Industrial Electronics}, vol.~56, no.~10, pp.
  3941--3950, 2009.

\bibitem{Steinfeld2006CommonInteraction}
A.~Steinfeld, T.~Fong, D.~Kaber, M.~Lewis, J.~Scholtz, A.~Schultz, and
  M.~Goodrich, ``{Common metrics for human-robot interaction},'' in \emph{HRI
  2006: Proceedings of the 2006 ACM Conference on Human-Robot Interaction},
  vol. 2006.\hskip 1em plus 0.5em minus 0.4em\relax Association for Computing
  Machinery, 2006, pp. 33--40.

\end{thebibliography}
